# Integrating Flow Theory and Adaptive Robot Roles: A Conceptual Model of Dynamic Robot Role Adaptation for the Enhanced Flow Experience in Long-term Multi-person Human-Robot Interactions

Huili Chen*
MIT Media Lab
hchen25@media.mit.edu

Sharifa Alghowinem
MIT Media Lab
sharifah@media.mit.edu

Cynthia Breazeal
MIT Media Lab
cynthiab@media.mit.edu

Hae Won Park
MIT Media Lab
haewon@media.mit.edu

## ABSTRACT
In this paper, we introduce a novel conceptual model for a robot's behavioral adaptation in its long-term interaction with humans, integrating dynamic robot role adaptation with principles of flow experience from psychology. This conceptualization introduces a hierarchical interaction objective grounded in the flow experience, serving as the overarching adaptation goal for the robot. This objective intertwines both cognitive and affective sub-objectives and incorporates individual and group-level human factors. The dynamic role adaptation approach is a cornerstone of our model, highlighting the robot's ability to fluidly adapt its support roles—from leader to follower—with the aim of maintaining equilibrium between activity challenge and user skill, thereby fostering the user's optimal flow experiences. Moreover, this work delves into a comprehensive exploration of the limitations and potential applications of our proposed conceptualization. Our model places a particular emphasis on the multi-person HRI paradigm, a dimension of HRI that is both under-explored and challenging. In doing so, we aspire to extend the applicability and relevance of our conceptualization within the HRI field, contributing to the future development of adaptive social robots capable of sustaining long-term interactions with humans.

**ACM Reference Format:**
Huili Chen*, Sharifa Alghowinem, Cynthia Breazeal, and Hae Won Park. 2024. Integrating Flow Theory and Adaptive Robot Roles: A Conceptual Model of Dynamic Robot Role Adaptation for the Enhanced Flow Experience in Long-term Multi-person Human-Robot Interactions. In *Proceedings of the 2024 ACM/IEEE International Conference on Human-Robot Interaction (HRI '24), March 11–14, 2024, Boulder, CO, USA.* ACM, New York, NY, USA, 11 pages. https://doi.org/10.1145/3610977.3634945

*Corresponding Author.
Acknowledgement: This work was supported by the Inclusive AI Literacy and Learning Gift Grant from DP World and the IITP Grant funded by the Korean government (MSIT)..

## 1 INTRODUCTION

Adaptivity is critical for social robots to interact successfully with humans, a notion underscored in previous Human-Robot Interaction (HRI) literature [47, 82]. It is well-established that adaptivity is integral for a robot or agent to be perceived as a team partner when collaborating with humans [42, 44]. Adaptive social robots typically exhibit capabilities such as understanding and expressing emotions, engaging in high-level dialogue, adjusting according to user responses, establishing social relationships, responding to varied social situations, and embodying different social characteristics and roles [26].

However, developing adaptive social robots for real-world settings continues to be a formidable challenge in HRI, as identified by Tapus et al. [82]. As argued by [11], achieving the right balance of autonomy, proactiveness, and user-driven interaction is essential for appropriate human-centered robot adaptation. Excessive autonomy and pro-activeness might lead to robots behaving "selfishly," seeming uncaring and detached from their human users. This equilibrium is particularly challenging due to its variability across different parameters such as individuals, HRI patterns, and cultural contexts. This challenge, compounded by the growing urgency to develop robot adaptivity, becomes even more pronounced as the context shifts from single-person HRIs (SHRIs) to multi-person HRIs (MHRIs).

In MHRIs, presetting a fixed optimal robot behavior policy becomes exponentially more difficult, considering that both group- and individual-related factors continuously shape group dynamics. Successful MHRIs, therefore, place a greater reliance on a robot's adaptability. With the rising prevalence of social robots in everyday life, robots are not just interacting with individuals but also engaging with multiple people, further emphasizing the need for research on robot adaptability in long-term interactions with human groups. This includes exploring how to design, develop, and evaluate long-term adaptive interactions between robots and humans in the wild.

In this work, a novel and context-independent conceptualization for designing long-term adaptive multimodal interactions between a social robot and a small human group is introduced. Drawing insights from foundational work within the HRI field and in psychology, this conceptualization introduces a hierarchical interaction objective grounded in the flow experience in psychology [18, 19], serving as the overarching adaptation goal for the robot. This objective intertwines both cognitive and affective sub-objectives and incorporates individual- and human-group-level factors.

Central to this conceptualization is the dynamic robot role-adaptation strategy, a pioneering approach aimed at guiding the adaptation of robot behavior along the support role. Both empirical work and theories pertaining to the robot role-adaptation suggest great promise of adapting a robot's behaviors along the robot's support role for effective continuous long-term adaptation. Beyond introducing concepts, this work discusses the limitations and potential applications of the proposed conceptualization, thereby illuminating key challenges and future trajectories for long-term HRI.

A notable focus of this work is on the largely under-explored MHRI paradigm, which, despite its increasing urgency and relevance, has been overshadowed by research on single-person interactions in domains such as education [6] and healthcare [41]. With an emerging body of HRI studies beginning to explore the design of a robot's interaction role and social behaviors in multi-person contexts [40, 83, 92], the under-development of conceptual frameworks and design principles for MHRI is evident. The accelerating integration of social robots into everyday human environments highlights the need for expanding research horizons to include interactions involving small human groups. Consequently, developing MHRI-oriented conceptualization is essential for advancing the HRI field and maximizing the potential benefits of social robots in everyday life.

Purposefully oriented towards the multi-person paradigm, this work aspires to extend the applicability and relevance of the proposed conceptualization to a broader spectrum within the HRI field. The aim is to address real-world long-term human-robot interactions in environments where robots are integral components of human ecosystems, anticipating a future where such interactions are increasingly prevalent due to advancements in generative artificial intelligence (AI).

Positing that single-person interaction is a special case within the multi-person paradigm, this work emphasizes the importance of long-term interactions across multiple sessions. The evolution of HRI research towards more intricate, in-the-wild, and long-term multi-person studies necessitates the design of robots capable of navigating these scenarios.

In conclusion, this proposed conceptualization offers several unique contributions:

- It provides a comprehensive overview of the state-of-the-art research on the design of robot roles in MHRI and their effects on the human groups.
- It marks an inaugural attempt to conceptualize an overarching adaptation goal for robot interaction, grounded in the flow experience, and incorporating both group- and individual-level design factors and considerations.
- The dynamic robot role adaptation approach is conceptualized and formalized for the first time in the HRI field, representing a pioneering effort in the literature for the MHRI context.
- The integration of two innovative designs into a cohesive robot adaptation framework is delineated, with an emphasis on their independent applicability.

These contributions lay the groundwork for practical advancements and provide a roadmap for developing adaptive social robots capable of engaging in long-term interactions with small human groups, thereby leading to the full realization of the potential benefits of social robots in everyday human life.

The following sections are structured to introduce each conceptualization stage: Goal Design of Robot Behavior Adaptation (Section 3), Robot Behavior Design (Section 4), and Dynamic Robot Role-adaptation (Section 5). Each section provides an overview and related works review followed by a detailed description of the proposed approach. In Section 6, we discuss the interconnections between the flow-experience-based adaptation objective and dynamic role-adaptation with other potential robot behavior and adaptation design approaches. We conclude with an examination of the limitations in our conceptualization and anticipated future directions for long-term MHRI adaptation research.

## 2 STATE-OF-THE-ART: BEHAVIORAL ADAPTATION OF SOCIAL ROBOTS

Robot adaptation strategies in SHRIs have been extensively examined, demonstrating various benefits. Social robots in this paradigm utilize diverse interaction modalities and user data for computational personalization of behaviors. Prior SHRI studies have leveraged activity-related data and user behavior logs for adaptation [5, 15, 49]. Social signals sensed from individual users, e.g., visual smiles and vocal laughs, have also been integrated into the robot's computational models for personalization, e.g., [32, 58, 81, 97]). This multimodal social personalization has been shown to foster positive human-robot interactions and relationships, e.g., rapport building and cooperation [46].

A variety of computational techniques, particularly focusing on Reinforcement Learning (RL), have been employed for multimodal social personalization in SHRI. These techniques integrate multimodal user and interaction data into RL-based models, guiding a robot's social-affective behaviors. For example, a robot, guided by an internal social motivation system that takes in user affective and interactive signals as input, learned how to select the behaviors that would maximize the pleasantness of the interaction for its peers [81]. Sensed multimodal user signals, e.g., vocal laughs and visual smiles, could be used as inputs for a reward function to analyze a user's manifested reactions and evaluate the robot's newly executed action in a RL-based adaptation model for optimized robot action selection, as shown in the prior long-term HRI studies [32, 58, 97]). Similarly, an RL behavior policy could take in a user's emotional state inferred from their nonverbal behavior along with the activity progress to personalize its social-affective behaviors, e.g., an empathetic chess companion robot picks different empathetic strategies and encourages a child playing chess in a personalized way [48]. In summary, the prior studies above demonstrated both benefits and sample implementations of a robot's behavioral adaptation in SHRIs.

In contrast, robot adaptation strategies in MHRIs remain underexplored. The majority of prior MHRI studies either directly teleoperated social robots or implemented simple rule-based models as social robots' behavior policies in interactions. A Wizard of Oz (WoZ) teleoperation framework has often been used to dismiss the robot control and autonomy challenges embedded in MHRIs, where a hidden human controls the behavior of the robot (e.g., [70]). The WoZ method has also been used as an interactive approach to examining the baseline effects of the robot's mediation methodology, e.g., [24, 94], as well as collecting realistic interaction data and building classification models, e.g., [56].

In other studies, rule-based models have been used as robot behavior policies in MHRI, (e.g., [12, 30, 71]) and this behavior policy is sometimes augmented with a perception system, e.g., [4, 12]. For example, guided by a rule-based behavior policy, a robot acted as a moderator in a human group interaction centered around a tablet-based assembly game to promote group cohesion and task performance [71], and collaborated with human pairs using a visual perception system in problem-solving cognitive tasks [12]. However, rule-based models do not allow for a robot's real-time behaviors to be adaptive.

Very few prior MHRI studies (e.g., [36]) have attempted to design, develop, or implement robot behavior adaptation policies. In Irfan et al. [36], an emotionally adaptable agent designed for long-term MHRI was able to model the real-time emotional state of users to dynamically adapt dialogue utterance selection using a Partially Observable Markov Decision Process (POMDP). To the best of our knowledge, other commonly used robot adaptation techniques in SHRIs have not been applied to MHRIs yet. Despite the under-exploration of robot adaptation in MHRI, the benefits of robot adaptation may likely even magnify in the MHRI context when compared with the SHRI context, as the interactions within human groups are inherently social and constantly dynamic, requiring a robot to understand group dynamics and processes to display appropriate behaviors in time.

## 2.1 Limitations in the State-of-the-Art

Overall, research into long-term adaptation in HRI remains a relatively underexplored field with several limitations. Firstly, although a diverse range of robot adaptation strategies have been introduced in SHRIs, the objectives of adaptation in many of these interactions have typically been single-dimensional and static throughout the entirety of the interaction. This method falls short of capturing the multifaceted nature of human interaction, which inherently consists of multiple goals and dimensions (e.g., cognitive and emotional) that adapt and evolve over time.

Second, adaptation design in many prior personalized HRIs studies has predominantly focused on content customization, overlooking the unique potential of social robots as embodied social agents. For example, personalization of the activity challenge was often achieved by adjusting the presented learning content or curriculum while keeping the robot's behavior policy static, as evidenced in [77]. This approach has, to some extent, underserved the full potential of robots as social agents, often reducing them to personalized content providers.

Moreover, the extent to which current robot adaptation strategies, primarily developed for SHRIs, can be directly generalized to multi-person and long-term interactions in real-world settings remains uncertain. The complexities arising from group interactions challenge existing SHRI theories and approaches [69]. In recent years, HRI researchers such as [38] have advocated for more interdisciplinary MHRI research, incorporating theories from social sciences. Among all the conceptual and empirical works in MHRI, there is a particularly noticeable gap in the exploration of interdisciplinary approaches and methods for robots to adaptively engage in multi-person settings.

Lastly, a generalized conceptualization for robot adaptation in long-term HRI is noticeably under-explored in the field. To our knowledge, no prior work proposed the integration of psychological theories to formulate a novel, generalized approach for designing long-term adaptive HRI, applicable to both single- and multi-person paradigms and across various contexts. Consequently, our work endeavors to address these identified limitations in the field.

## 3 DESIGN: ROBOT'S ADAPTATION GOAL

The initial step in designing robot adaptation involves conceptualizing a well-defined group interaction objective. In human-human interactions (HHI), the presence of a group goal is paramount, significantly influencing group effectiveness [61, 98]. The empirical significance of group commitment to these goals has been well-documented in preceding HHI studies [3, 93]. Viewing through the lens of a robot computational policy as presented in Section 2, establishing a group interaction objective becomes instrumental in shaping how a robot engages with and adapts its behavior to assist a human group in reaching this objective. To articulate this overarching goal of the robot's adaptation, we leverage the theory of flow experience from psychology to design a hierarchical interaction objective.

## 3.1 Related Work: Flow Theory

The concept of flow experience was initially formulated to characterize an individual's positive subjective experiences (mental states) within a task context [18, 19]. As illustrated in Fig. 1a, an optimal experience emerges when there is a balance between an individual's skill levels and the challenges presented by an activity. A suboptimal experience might result from either an excess or a deficiency of challenges, leading to feelings of boredom or anxiety, respectively [59]. Subsequent research has identified that flow experiences, or optimal experiences, can also emanate from diverse social interactions such as cooperative, interpersonal, conversational, and sport activities [25, 51, 95]. Notably, social flow experiences have been observed to be potentially more intense and enjoyable compared to solitary flow [51, 95]. For an in-depth exploration of various concepts of flow in social contexts, refer to [95]. Given its potential to enhance positive interaction effects, the concept of flow experience is integrated into our proposed hierarchical objective, especially in social MHRI contexts.

Optimal experience in social interaction is multidimensional, encompassing both cognitive and affective aspects [85]. The cognitive dimension is manifested in users' skill levels, knowledge, and performance, while the affective dimension is centered around users' affective experience. The latter could be assessed through rapport, indicative of a harmonious connection and a meaningful experience [18]. The user's skill-challenge balance not only influences their feelings, but the affective dimension can also significantly impact the cognitive one. The presence of rapport within a group fosters a conducive environment for accomplishing challenging tasks necessitating mutual commitment [86], and its effectiveness has been validated in various contexts, such as peer-tutoring [75] and negotiation [21]. Therefore, both the affective and cognitive dimensions co-regulate each other.

In the vein of an individual's optimal experience, group flow was conceived inspired by individual flow experience, aiming to depict a unique kind of flow occurring in group settings. It is a collective phenomenon, qualitatively distinct from individual flow and likely surpasses a mere aggregation of individuals' flow experiences in a group [60]. Several conditions unique to group flow have been identified, including equal participation, familiarity within the group, and effective communication [67].

## 3.2 Proposed Approach

As illustrated in Fig. 1b, we define the overarching goal of a robot's adaptation as enhancing user flow experience, drawing inspiration from the definitions of flow experience or optimal experience. This objective extends to improving an individual's flow experience in a single-person interaction context, and the group's flow experience in a multi-person context. Fig.1c illustrates the various affective states users may experience, contingent on their task challenge and skill level. Conversely, Fig.1d demonstrates how robot behaviors can potentially guide a user back to the optimal flow experience when they deviate from the zone.

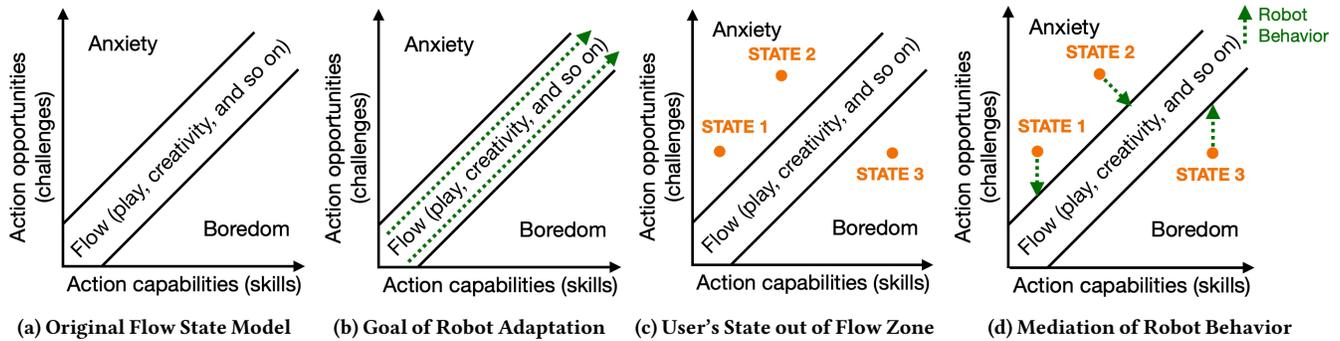

Figure 1: This set of figures illustrates how the design of the robot's adaptation goal is grounded in the flow experience and how a robot might positively mediate the flow state of an individual or a group over time via the dynamic role adaptation approach. (a) depicts the original model of the flow state as presented in [55], which has both cognitive dimension (e.g., task challenge and user skill) and affective dimension (e.g., anxiety, boredom, flow). (b) illustrates that the goal of the robot's adaptation is to guide users to remain in the flow zone and improve their skills over time. (c) showcases three user states that are out of the flow zone, with an orange dot representing the challenge-skill state users are in at a specific time. State 1 and State 2 represent anxiety states, where perceived challenges surpass skills, whereas, in State 3, boredom emerges as skills exceed the perceived challenges of the activity. (d) further illustrates how a robot can intervene to realign a user with the flow state. The dynamic role-adaptation approach allows the robot to exhibit varied role behaviors, adjust perceived challenge levels, and enhance skills with the aim of maintaining a balance between challenges and skills, thereby assisting individuals in remaining in the flow state and improving their skills over time. A green dashed arrow signifies the impact of robot behavior on the user's challenge-skill state, and the direction of an arrow indicates how a specific robot behavior delivers this impact. When users find themselves in State 1 or State 2, the robot can exhibit supportive behaviors to guide them back into the flow zone. The variance in arrow direction between State 1 and State 2 suggests different supportive behaviors aimed at modifying the challenge level and/or enhancing skills to varying degrees, such as offering direct assistance or providing step-by-step hints. For users in State 3, the robot can introduce additional challenges or pose stimulating questions to realign them with the flow zone.

Given that the flow experience encompasses both cognitive and affective dimensions, we further delineate the hierarchical objective of robot adaptation into two sub-objectives: fostering cognitive and affective dimensions at both individual and group levels, considering the difference in psychological constructs between group and individual flow. This conceptualization ensures that both cognitive-affective and group-individual dimensions are pivotal in guiding a robot's behavior policy, as illustrated in Fig. 2. Subsequent subsections will delve into the cognitive and affective sub-objectives and explore the compositional distribution of the four sub-objectives comprising the overall adaptation goal.

**Clarifications:** When designing group-level sub-objectives, we focus on the human group interacting with the robot and exclude the robot from the group to avoid confusion.

*3.2.1 Cognitive Sub-objectives.* When a group's interactive capacity or skill doesn't align with the activity's challenge, the robot can intervene. For instance, if the task is overly challenging, the robot can scaffold the group; if too simple, it can introduce more challenging prompts. A group's interactive capacity is evident at both individual and group levels, reflecting in **group performance** and **interaction resource distribution**. These elements guide the robot in determining which member(s) or the entire group to support at any given time. A key difference between SHRI and MHRI is that in MHRI, decisions about resource allocation among individuals are additional considerations. Equitable allocation of interaction resources in both HHI and MHRI is vital for collaboration and individual cognitive growth [39, 90]. Effective collaboration in HHI, marked by a balanced pattern of arguments, fosters cognitive growth through socio-cognitive conflict [90]. Similarly, equal allocation in MHRI leads to a more positive perception of group relations [39].

When designing a robot's behavior adaptation policy, taking into account both the group's social dynamics and the social consequences of the robot's resource distribution has been advocated by prior MHRI work [63]. Other MHRI work concurs, suggesting a robot's role in a team is twofold: managing interaction resources and optimizing group performance [71, 72]. A robot behavior policy solely optimizing team performance may inadvertently foster intergroup bias by favoring high performers [63]. These prior works altogether indicate that a robot behavior policy that maximizes the group performance at the cost of discriminating low performers or unequally distributing resources would not likely help the group achieve the optimal experience. Awareness of both group performance and resource allocation is essential for fostering inclusive and high-performing MHRI interactions.

*3.2.2 Affective Sub-objectives.* Designing affective sub-objectives necessitates considering both **affective states of individuals** and **group-level interpersonal dynamics**. Individual emotions and mental states influence reasoning, attention, learning [91], and overall group dynamics [31]. Constructs such as valence, arousal, and engagement, which have been widely employed in HRI studies, could be used to capture an individual's affective states. The valence-arousal scale, a widely-used measure, captures a user's pleasure and excitation levels [65]. Engagement, a measure of the quality of connection during interaction [66, 74], is also associated with valence and arousal in various contexts [10, 64]. However, the relevance and manifestation of these constructs vary across contexts. For instance, focused attention is critical in educational settings [68, 78], while excitement is key in entertainment [8]. When designing the individual-level affective sub-objective, selecting and contextualizing these affect constructs is thereby crucial.

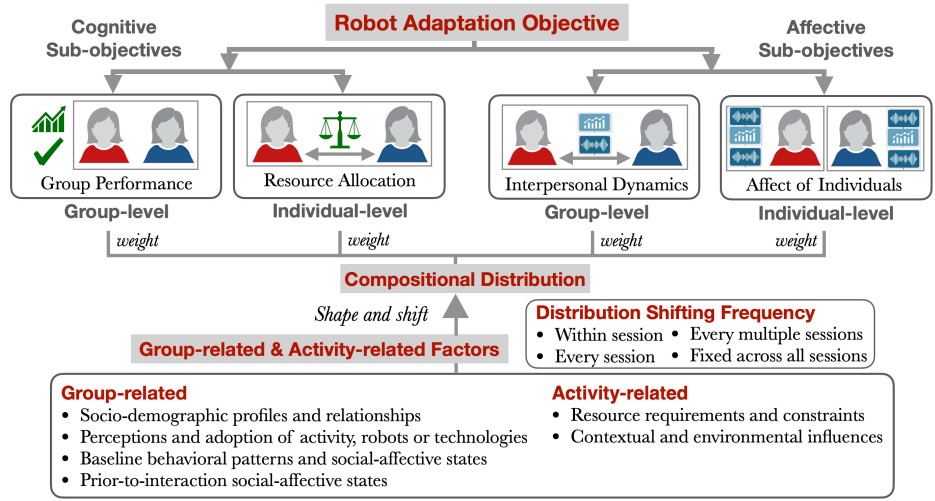

**Figure 2:** Rooted in the concept of flow experience, the hierarchical robot adaptation objective encompasses both cognitive and affective sub-objectives. These are further divided into group-level and individual-level objectives. The compositional distribution of these four sub-objectives are context-dependent, influenced by a myriad of group and activity-related factors. During long-term interactions, this compositional distribution is dynamic and apt to evolve over time with the goal of guiding users towards the flow zone in an upward direction over time. The frequency of these shifts in compositional distribution can vary—occurring within sessions, across multiple sessions, or remaining fixed—depending on the nature of the interaction and the users' profiles.

An affective sub-objective to assess the group-level interpersonal dynamics is also essential. A group is more than the sum of its parts [1, 27], necessitating a holistic approach to understanding group dynamics. Simply aggregating the group members' affective states to infer the group-level affective dynamics dismisses the mutual regulation of each other's affect among the group members, and may not sufficiently capture the nuance and complexity of the group's dynamics [31]. To capture more nuanced dynamics among people in a group, psychological constructs exclusively designed to understand the within-group affective dynamics are important, e.g., behavior-based dyadic rapport [87, 88], interpersonal synchrony [53], and group cohesion and structure [9, 27, 31]. For example, the dyadic rapport could be inferred by measuring the dyad's mutual attentiveness, positivity and interpersonal coordination [87, 88].

Using both individual- and group-level affective constructs, the affective sub-objectives provide a more holistic, multifaceted view of the human group. It is important to notice that individual-level affective states and group-level interpersonal dynamics are often interdependent [31, 84]. In a MHRI context, when the affective states of individuals in the group mismatch, the robot's adaptation policy may need to determine which person(s) its behavior targets to resolve the mismatch. By addressing an individual's affect, the robot's actions could also affect the group's dynamics, and vice versa.

*3.2.3 Compositional Distribution of Sub-objectives.* In long-term interactions, the goal of the robot is to scaffold users, guiding them to ascend in the flow zone over time. The optimal compositional distribution that enables the robot to efficiently achieve this goal may evolve. Considering the constraints of available interaction resources in a session, such as total interaction time and opportunities for displaying robot behavior, a robot's adaptation policy might sometimes face a dilemma when choosing between sub-objectives. Concentrating solely on one aspect of the interaction may not yield an optimal experience in the long term, as different sub-objectives might regulate or covary with each other. Determining an appropriate compositional distribution of all sub-objectives is pivotal for optimizing the overall interaction experience effectively. The sub-objectives don't necessarily contribute equally to the overall interaction objective. This distribution is contingent upon specific interaction scenarios and human groups.

Consequently, identifying key group-related and activity-related factors that significantly shape the distribution is essential. Group-related factors might include *(d.1)* a group's sociodemographic profiles, characteristics, and relationships, such as age, temperament, and personality traits; *(d.2)* their perceptions and adoption of the activity, robots, and technologies, including their openness to robot interaction and perceived robot social attributes; *(d.3)* their baseline behavioral patterns and social-affective states in group interactions before introducing robot interaction, for example, conversational turn-taking and engagement levels; and *(d.4)* their social-affective states right before a new robot session starts, such as interpersonal conflict and fatigue. Activity-related factors might comprise *(a.1)* activity status and progress, such as the number of completed sessions, activity duration, and changes in activity schedule; *(a.2)* resource requirements and constraints, such as physical or virtual space, and activity equipment; and *(a.3)* contextual and environmental influences, including location, background noise, and weather. All these factors collectively shape the interaction objective's compositional distribution.

While the initial distribution may be static in some studies, it may shift over time in response to changes in group and activity factors. The goal of group interaction evolves through feedback and individual adaptation [23, 67], necessitating adjustments in distribution. Such shifts can occur at varying frequencies, driven by dynamic factors that are likely to vary within and between sessions. Recognizing these shifts is crucial for optimizing interaction. For instance, *(d.4)*, *(a.1)*, and *(a.3)* are all dynamic and likely to vary from session to session or even within each session, thus driving the overall hierarchical interaction objective to shift across each activity session if identified as key factors shaping the sub-objective distribution.

## 4 DESIGN: ROBOT ROLE-BASED BEHAVIOR

In this section, we focus on the robot behavior design in our conceptualization of the long-term robot adaptation, presenting a structured approach based on the robot support role. While individual studies in the HRI literature have empirically explored this topic on the linage between robot roles and role-associated benefits in users, ours is the first attempt to formalize it as a holistic, generalizable design spectrum for robot behavior.

### 4.1 Related Work: Robot Support Role

A robot's support role profoundly affects human-robot interactions in both SHRI and MHRI. Since SHRI has been more extensively researched than MHRI, our emphasis here is on MHRI.

In MHRI, social robots can assume various roles, such as mediator, coach, or listener, depending on the interaction context. As per a recent MHRI survey by Sebo and colleagues [69], robot roles can be broadly classified based on activity-related support levels into three types: follower, peer, or leader. Robots can engage as a **leader**, offering explicit activity-related support as a coach, resource provider, or activity host. Examples include counseling couples [92], providing guidance in public spaces [28], or hosting group games [99].

Contrastingly, robots can also assume a **follower or learner role**, responding to interaction initiatives or following human instructions, aiming to subtly support multi-person interactions. A robot named *Kip1*, designed as a conversation companion, encourages non-aggressive dialogue between people through gesture cues [34]. Similarly, robots making vulnerable statements can positively influence group dynamics [89], and displaying emotions can enhance human perception of the robot [52].

A **peer-like robot role** involves robots initiating and leading interactions similarly to humans. Some robot roles, like the moderator, sit between the leader and follower roles on the knowledge spectrum. In the capacity of a mediator, robots can foster positive human group dynamics and even resolve conflicts, as demonstrated in multiple studies [7, 22, 70]. Acting as a moderator, robots can bolster a human group's social dynamics [73] and improve collaboration by posing questions [80].

As shown above, social robots can perform a plethora of support roles, which could be generally categorized on a spectrum from leader to follower based on the level of competence- or task-related support they can provide. MHRI research also reveals that the choice of role can have differential impacts on users. For instance, leader-like robots can assist groups in completing activities efficiently [28], while follower and peer-like roles can positively shape group dynamics, fostering trust and promoting harmonious conversations [7, 34].

For SHRI, this role design spectrum is equally applicable, and distinct benefits associated with each role are elaborated upon in related works such as [15].

### 4.2 Proposed Approach

Utilizing this support role spectrum, as illustrated in Fig. 3, is instrumental in developing robot behavior and subsequent adaptation. Role-based robot behavior design can facilitate the robot's integration into human interactions, as role differentiation is a fundamental aspect of human-human interaction [2, 76].

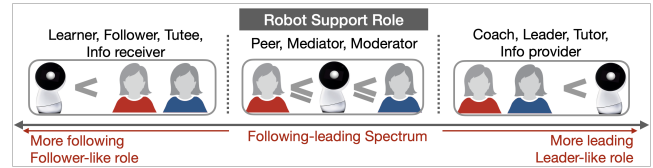

**Figure 3: This spectrum illustrates a variety of a robot's support roles based on the level of competence- or task-related support it can provide. It can assume the role of a follower, learner, or tutee within a team; conversely, it can adopt a more leading role in a human-robot group, serving as a coach, leader, or tutor. This spectrum can be either continuous or categorical. Previous research in both single- and multi-person HRI has demonstrated the benefits of roles at either end of the spectrum, encompassing both leader-like and follower-like roles.**

While Sebo and colleagues [69] initially categorized robot roles in MHRI into three distinct categories (follower, peer, leader), we propose a more fluid classification. We introduce a robot role spectrum, ranging from the most leading to the most following role, aiming to encapsulate the robot's support level relative to individuals in the human group.

Several design considerations are important when developing role-based behaviors for robots in MHRI. Firstly, understanding role differentiation in a human group is vital, as it is integral to human collaboration [2] and varies across different group sizes and contexts [17, 76]. Role differentiation needs to be assessed in designing MHRI, considering its prevalence and significance in human groups. A robot's support role is thus defined in relation to the role differentiation of human partners on the leader-follower spectrum. A robot assumes a more leader-like support role, such as coach or tutor, only if the support it offers surpasses that of all human group members. Conversely, a follower-like robot responds to or follows all individuals in the group. Other roles, like moderator or peer, are positioned between the leading and following persons, with exact positions dependent on specific activities and group dynamics.

Secondly, it is crucial to analyze the distinct benefits each robot role might offer for optimizing MHRI. Previous studies comparing robot support roles have demonstrated varied impacts on cognitive performance, affective states, social perceptions, and group dynamics [12, 15, 16, 33]. For example, studies have found differing effects on learning performance, affective expressivity, and blame attribution in SHRIs [15, 16, 33]. In another study, a cognitively reliable robot aided task performance, while a less reliable one facilitated explanations between people [12]. Evaluating the potential benefits of each robot support role for both individual and group interactions is essential in developing effective MHRIs.

## 5 DESIGN: DYNAMIC ROBOT ROLE-ADAPTATION

Assigning a robot's leader-follower role in advance is not always feasible, particularly in dynamic environments or during long-term interactions. In this section, we formalize dynamic robot role-adaptation as a method to guide real-time adaptations in a robot's support role during an interaction. The aim is to find an optimal balance and maximize the overall benefits of different robot roles. We also discuss the integration of this dynamic role adaptation with the hierarchical adaptation objective introduced earlier.

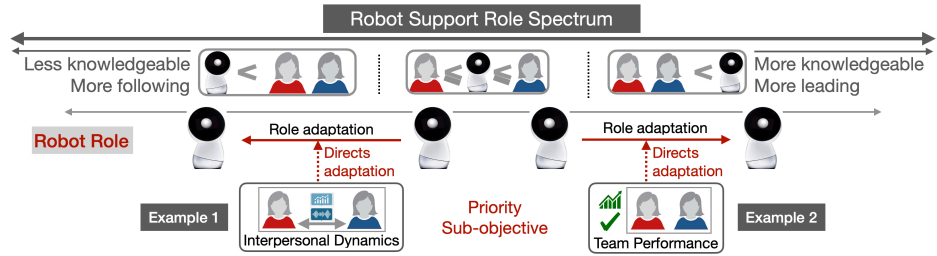

Figure 4: To harness the unique benefits of varying robot roles and optimize the robot's overall positive influence on the group or individuals, the robot adapts its role along the leading-following support role spectrum as the compositional distribution of its hierarchical adaptation objectives evolves over time, either within or across interaction sessions. The examples provided illustrates how the robot's role dynamically adjusts in response to changes in the priority sub-objective driving the adaptation. The estimated benefits of specific robot roles in achieving sub-objectives are grounded in empirical HRI studies; for instance, an emotionally vulnerable robot in MHRI has been shown to enhance group dynamics [89]. Example 1: When interpersonal dynamics are predominant in the hierarchical adaptation objective, the robot adopts a follower-like role to prioritize the affective aspect of the interaction, positively influencing the group's dynamics. Example 2: When team performance emerges as the priority among the sub-objectives, the robot assumes a more leading role, offering direct guidance and support to aid in task completion and enhance the team's overall performance.

## 5.1 Related Work: Adaptive Robot Roles

Empirical studies have demonstrated that enabling a robot to adaptively switch its role is more advantageous than static and predefined role assignment mechanisms in SHRI contexts, covering both physical cooperation [37, 50, 54] and social interactions [15, 35, 62]. For instance, a robot can adapt its leader-follower role in real-time, based on the human partner's intention, enabling more fluid collaboration in physical cooperation [37]. A study by Li et al.[50] designed an adaptive scheme to parameterize a spectrum of mixed robot roles, allowing a robot to decide when to complete the task independently and when to follow the human. Another study[54] implemented dynamic robot role allocation schemes for a cooperative load transport task, underscoring the importance of a robot's ability to adjust its role for enhancing its utility in cooperation.

Compared to physical human-robot cooperation, the paradigm of social interaction with role-adaptive robots is less explored but has already shown promise. A role-adaptive robot, for example, can enhance students' intrinsic motivation to learn creative dance by switching roles [62] or aid in learning Computer Science concepts by adaptively tracking and shifting task initiative [35]. Furthermore, such a robot can effectively promote learning and affective experiences by leveraging the benefits of both tutor and tutee roles [15].

While previous SHRI works have designed robots capable of adaptively switching roles during human interaction [15], formalizing this capability as an adaptation approach has yet to be done. In most current MHRIs, a robot's role is predetermined and remains constant throughout interactions, e.g.,[29, 79]. The study most aligned with using the role-adaptation approach in MHRIs involved collaboration between drones and human teams[45]. This study modeled the emergent dynamics in human-robot teams to guide leading-following behaviors, illustrating the potential of applying dynamic role-adaptation in MHRIs, even if the collaboration was physical rather than social.

In conclusion, given its proven advantages in SHRIs, the dynamic role-adaptation approach has substantial potential for generalization and application across both SHRI and MHRI contexts.

## 5.2 Proposed Approach

From the perspective of flow theory, our approach allows a robot to offer adaptive support, dynamically aligning the challenge of an activity with a user's competence level to sustain their engagement in the flow zone. Unlike several previous works, where the challenge was mainly adjusted through modifying learning content or curriculum while maintaining static robot behavior, e.g., [77], our approach fosters a personalized activity challenge by altering robot role-related behaviors. This involves the robot as an integral component of the dynamic environment with which a person interacts, diversifying engagement strategies and enhancing the promotion of non-cognitive goals like affective experiences.

For instance, should a user exhibit high competence in a given activity, the robot, acting as a follower or learner, could minimize task-related support and introduce playful and affective behaviors to elevate the group's positive engagement. As depicted in Fig.1, this method maintains a **dynamic balance** between nurturing action capabilities (skills) and modulating action opportunities (challenges), a fundamental aspect of the flow experience[19, 20, 96].

Given the distinct benefits of leader-like and follower-like roles, dynamic role-adaptation can harness these role-associated benefits to fulfill adaptation objectives, which vary across different groups and contexts. Different robot roles can uniquely guide users towards the flow zone, leveraging the distinct benefits of each role, as represented by the directional arrows in the color green in Fig. 1d. This flexibility provided by dynamic role-adaptation is essential for maintaining users in the flow zone throughout extended interactions.

In the context of MHRIs, the potential benefits of dynamic role adaptation could be even more pronounced due to their inherently social nature compared to SHRIs. A robot's capability to positively influence group dynamics can enhance the group's social flow, resulting in a more enjoyable and intensified flow experience [51, 95]. Similar to human-human interactions where partners can heighten each other's immersion in the flow zone, a robot can influence group members reciprocally to meet individual flow conditions.

Implementation of this approach necessitates several **design considerations**. Firstly, the compositional distribution of the sub-objectives in the hierarchical adaptation goal directs the robot's adaptation on the leader-follower spectrum. As shown in the example 2 illustrated in Fig. 4, a robot could adapt to a more leader-like role if its group performance sub-objective overweighs the other sub-objectives. If its affective sub-objectives become priorities for the robot to achieve as shown in example 1 (Fig. 4), it could adapt from the *leader-like* role to a *mediator* or *peer* role that offers encouragement and acknowledgment to help reallocate resources among members and promote positive group dynamics.

Furthermore, defining the role-adaptation frequency is also crucial for effective adaptation. In prior SHRIs, robots adapted their support roles at different frequency levels, e.g., shifting its role multiple times within each interaction session [15], and only once per session [62]. The adaptation frequency needs to be contextualized in each interaction context to ensure that the robot's behavior remain fluid and consistent, as well as its social characteristics remain positive from the users' perspective. An inappropriate frequency may lead to a robot's ineffective personalization or and negative user experiences. For example, a sudden, over-frequent shift from a leader-like demonstrator role to a vulnerable help-seeker role in the same activity may confuse the group with why the robot seems to suddenly forget the knowledge it already masters.

Finally, it is critical to distinguish between robot support role and robot social role, identity, or personality. The dynamic role adaptation approach pertains only to the former; altering the latter could introduce confounding variables. Social attributes and personality can significantly impact user interaction, perception, attitude, and social relationship with the robot, e.g., [43, 57], thereby adding complexity to MHRI design and evaluation. Thus, the adaptation mechanism solely guides specific supportive behaviors exhibited by the robot during interaction, e.g., offering help as a leader, making mistakes as a learner.

## 6 DISCUSSION ON FUTURE DEVELOPMENT

In this work, our focus has been on the robot's leading-following support role as a critical dimension in robot behavior design, especially in enhancing the user flow experience. It's noteworthy that variations of the robot support role spectrum likely exist and can be applied to role design. For instance, an affective-cognitive robot role spectrum might indicate whether a robot's support is aimed at fostering the cognitive or affective aspect of the interaction. Similarly, a motivational-informational role spectrum could enable a robot to offer either more information-oriented or motivation-oriented support to a human group. Depending on the specific MHRI contexts, other variations might be more suitable than the general following-leading spectrum.

It is essential to clarify that we do not advocate for the robot's support role as the sole or primary axis for robot behavior design. Other axes could significantly facilitate a user's flow experience and warrant further exploration at the intersections of existing literature and psychological theories. While our approach is well-aligned with flow theory and uncovers potential synergies, it is not the only path; other potential adaptation approaches could yield synergistic effects when integrated with our proposed hierarchical adaptation goal, grounded in flow experience. In this regard, the dynamic role-adaptation approach is designed to complement, rather than supersede, other methods rooted in psychological frameworks for robot behavior design and adaptation.

Therefore, the development and utilization of alternative adaptation strategies are highly encouraged. Indeed, these alternative strategies might also enable users to achieve the hierarchical goals outlined in this study, underscoring the flexibility of the framework. Moreover, when combined with other robot adaptation design approaches, the role-adaptation approach could reveal additional synergies. For instance, human-robot nonverbal mimicry could be developed as an orthogonal design axis for robot behavior adaptation, enabling a robot to dynamically adjust its communication style in tandem with its support role, thereby delivering personalized interaction to individual users or human groups.

Our work serves as an initial step in the formulation of a comprehensive framework for facilitating user flow experiences through the robot dynamic role adaptation. This study provides a conceptual scaffold for designing adaptive robot behavior by integrating the principles of flow experience.

Building on strong evidence and past literature, our work makes significant conceptual contribution. Notably, it has been successfully applied in an empirical study on multi-human-robot interaction in the context of parent-child story-reading [13, 14], confirming its potential utility. Nonetheless, we encourage more empirical validation through more user studies. Future research could focus on case studies to validate the robustness, applicability, and generalizability of this framework. Future quantitative analyses will offer a more comprehensive validation of the framework.

However, this paper does not offer ready-to-use metrics or measurements for the flow experience in MHRI. We believe that the efficacy of robot adaptation should be evaluated within the context of specific interaction scenarios, thus leaving room for future research to operationalize this framework into measurable metrics. It should be also noted that our conceptual model differs from a ready-to-use practical design tool or system development framework for HRI. The current study aims to illuminate new research directions instead. Consequently, building upon our work, future research can focus on developing a more hands-on framework that guides HRI practitioners from ideation to system deployment, coupled with quantitative metrics for evaluation.

As for the implementation challenges, we acknowledge a potential technical gap between our conceptual formalization and the current state-of-the-art in autonomous HRI systems. Our proposed approach implies complex sensing and perception capabilities for the robot, particularly in multi-person interaction contexts. Such capabilities, although conceptually outlined, may pose technical difficulties given the current state-of-the-art. Hence, our conceptual work highlights the urgency and importance of developing robot perception systems capable of nuanced understanding of human interaction dynamics, emotions, and performance assessments.

In summary, our conceptual model sets the stage for future research endeavors to bridge the divide between theoretical insights and real-world applications. Extending our work, the future work could include more focused empirical studies, the design of precise metrics and measurement methods, the development of practical design tools or system frameworks, and the actualization of advanced robotic perception systems.

## 7 CONCLUSION

To our best knowledge, our work represents the first conceptual design approach for a robot's behavioral adaptation in the long-term HRI, being generalizable across both single- and multi-person interaction scenarios and applicable across diverse application domains. Our conceptualization is grounded in the flow experience principles from psychology. Central to our conceptual model is the dynamic robot role adaptation, facilitating a robot to adapt its support role—from leader to follower—to consistently maintain a dynamic balance between activity challenges and user skills. With the overarching aim of guiding users toward the optimal flow experience, the dynamic role-adaptation approach directs the robot to engage adaptively with human groups.